\documentclass[times,twocolumn,final]{elsarticle}

\usepackage{cag}
\usepackage{framed,multirow}

\usepackage{amssymb}
\usepackage{latexsym}

\usepackage[hyphens]{url}
\usepackage{xcolor}
\usepackage{subfig}
\definecolor{newcolor}{rgb}{.8,.349,.1}

\usepackage{booktabs}
\usepackage{xspace}
\usepackage{soul}

\newcommand{\removed}[1]{}
\newcommand{\added}[1]{#1}

\newcommand{\camreadyadd}[1]{#1}

\usepackage{hyperref}

\usepackage[switch,pagewise]{lineno} 

\journal{Computers \& Graphics}

\begin{document}

\verso{Preprint Submitted for review}

\begin{frontmatter}

\title{Deep traffic light detection by overlaying synthetic context on arbitrary natural images}%


\author[ufes]{Jean Pablo \snm{Vieira de Mello}\corref{cor1}\cortext[cor1]{Corresponding author.}}
\ead{jeanpvmello@gmail.com}
    
\author[ufes]{Lucas \snm{Tabelini}}

\author[ufes]{Rodrigo \snm{F. Berriel}}

\author[ufes,ifes]{Thiago \snm{M. Paixão}}

\author[ufes]{Alberto \snm{F. de Souza}}

\author[ufes]{Claudine \snm{Badue}}

\author[trento]{Nicu \snm{Sebe}}

\author[ufes]{Thiago \snm{Oliveira-Santos}}

\address[ufes]{Universidade Federal do Espírito Santo, Brazil}
\address[ifes]{Instituto Federal do Espírito Santo, Brazil}
\address[trento]{University of Trento, Italy}

\address[1]{Address, City, Postcode, Country}
\address[2]{Address, City, Postcode, Country}

\received{\today}

\tnotetext[tnote1]{Article DOI: \url{10.1016/j.cag.2020.09.012}.}

\begin{abstract}
Deep neural networks come as an effective solution to many problems associated with autonomous driving. By providing real image samples with traffic context to the network, the model learns to detect and classify elements of interest, such as pedestrians, traffic signs, and traffic lights. However, acquiring and annotating real data can be extremely costly in terms of time and effort. In this context, we propose a method to generate artificial traffic-related training data for deep traffic light detectors. This data is generated using basic non-realistic computer graphics to blend fake traffic scenes on top of arbitrary image backgrounds that are not related to the traffic domain. Thus, a large amount of training data can be generated without annotation efforts. Furthermore, it also tackles the intrinsic data imbalance problem in traffic light datasets, caused mainly by the low amount of samples of the yellow state. Experiments show that it is possible to achieve results comparable to those obtained with real training data from the problem domain, yielding an average mAP and an average F1-score which are each nearly 4 p.p. higher than the respective metrics obtained with a real-world reference model.
\end{abstract}

\begin{keyword}
\KWD Traffic light\sep Synthetic data\sep Deep learning\sep Object detection\sep Image context
\end{keyword}

\end{frontmatter}


\section{Introduction}

Autonomous driving has been a very active research area in the past few years~\cite{badue2019selfdriving,DBLP:conf/ijcnn/BerrielTCGBSO18,berriel2017cag, mutz_large-scale_2016, lyrio_image-based_2015, berriel_ego-lane_2017}. For an autonomous vehicle to be safe, it has to be aware of its surroundings, which includes detecting pedestrians, traffic signs, and traffic lights. In this work, the focus is on traffic light detection. The goal of traffic light detection is to localize (with a bounding box) each traffic light in an input image and recognize its state (e.g., green, yellow or red). The accurate detection of traffic lights is essential for autonomous vehicles that are intended to travel on public streets, otherwise, the chances of an accident rise considerably~\cite{tl_running}. Hence, there have been numerous works tackling this problem.

Most traffic lights follow a similar pattern: three bulbs (one for each state) in a black case~\cite{jensen2016vision}. Because of this pattern, the first methods proposed for traffic light detection relied on hand-crafted feature engineering. Those features were designed mainly based on colors~\cite{diaz2015robust,gomez2014traffic} and shapes~\cite{trehard2014tracking}. Nevertheless, this approach has limited robustness, since hand-crafted features tend to overfit. To increase robustness and generalization, a learning-based approach may be used, such as SVM~\cite{jang2014multiple}, AdaBoost~\cite{gong2010recognition}, or JointBoost~\cite{haltakov2015semantic}. In particular, deep neural networks (DNNs) have gained traction in recent years, outperforming traditional methods in an end-to-end manner~\cite{jensen2017evaluating}.

\begin{figure*}[t]
    \begin{center}
       \includegraphics[width=\linewidth]{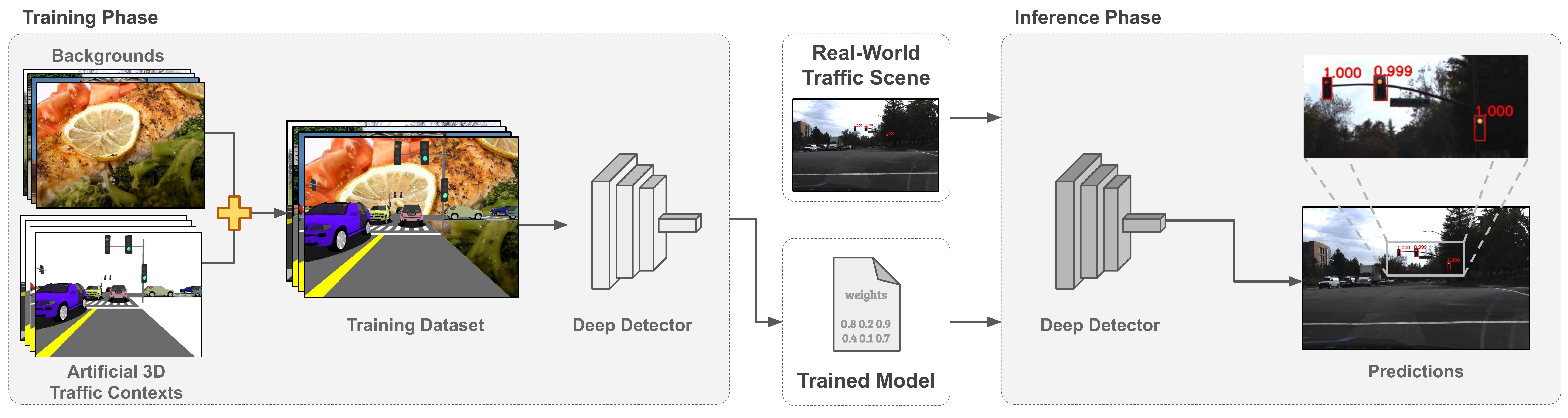}
    \end{center}
    \caption{Overview of the proposed method. Artificial traffic scenes are blended in natural backgrounds, composing a dataset to train a deep detector. Then, the trained model is ready to detect and recognize real traffic lights.}
    \label{fig:overview}
\end{figure*}

DNNs have been applied to many problems in autonomous driving, such as the detection of traffic signs~\cite{tabelini2019ijcnn} and pedestrians~\cite{ouyang2013joint,sarcinelli2019cag}. One of the reasons is that state-of-the-art deep detectors (e.g., Faster R-CNN~\cite{ren2015faster} and YOLO~\cite{redmon2016you}) can be applied to different tasks with little tuning. Nonetheless, their use for the traffic light detection task has been only recently enabled, with the release of large public datasets containing traffic light annotations~\cite{gonzalez2018annotated,fregin2018driveu}.
This highlights one of the main issues with deep learning: the need for large amounts of annotated data. The processes of acquiring and annotating data for the target domain (i.e., the one in which the application is expected to operate) is usually very expensive, requiring several working hours (in this case, for image acquisition and manual annotation, conducted by humans). This is even more critical for the detection task, since each traffic light in the image has to be marked with a bounding box, along with its state. \added{Some works try to alleviate the need for data with the use of generative adversarial networks (GANs)~\cite{arruda_cross-domain_2019}; however, it usually requires some level of annotation. For example, in the case of using GANs to generate artificial traffic scenes (backgrounds) without traffic lights, it would be required training images of the environment with weak annotations indicating the presence/absence of traffic lights. Otherwise, the network could inappropriately learn to generate traffic lights as background.} Moreover, traffic light datasets are typically imbalanced, given that yellow state is less likely to be found. Since most networks struggle with imbalanced data, extra effort on handling data distribution between the classes of interest is usually necessary.

To address the class imbalance and annotation effort issues in the context of traffic sign detection, Torres \textit{et al.}~\cite{tabelini2019ijcnn} proposed to combine arbitrary non-traffic-related natural images as background with templates (i.e., pictograms) of traffic signs to train a deep detector.
This work aims to adapt the concept introduced by Tabelini \textit{et al.}~\cite{tabelini2019ijcnn} to the traffic light domain, which has its own set of challenges (e.g., vehicles' tail-lights and other lights, in general, can hinder detection).
Therefore, this work also investigates the effect of blending non-realistic synthetically generated samples (using low-quality computer graphics) of traffic context on the natural backgrounds to enhance the detection performance. To evaluate the proposed method, experiments were performed using several traffic light databases from the literature. Experimental results showed that an adapted application of the method \cite{tabelini2019ijcnn} (using templates) yields low results (average mAP of 26.78\%). However, they showed that the method can be further enhanced by replacing the templates by the synthetic context, yielding an average mAP of 50.08\%, 4 p.p. higher than the obtained with real-world data training. These results indicate the feasibility of training deep neural networks without real-world samples from the target domain, which corroborates the results of \cite{tabelini2019ijcnn}, and that adding low-quality context to the training image backgrounds can improve the results even further. Moreover, the results also show that the method can boost (by more than 11 p.p. of mAP, in average) the results obtained with real-world data training when using the proposed method to augment the training data.

\section{Related works}
The literature on traffic light detection is roughly categorized into model- and learning-based methods. Most of the first works used model-based methods, which usually rely on features such as the color~\cite{li2018traffic} and shape~\cite{gomez2014traffic} of traffic lights. However, most recent works propose using learning-based methods, which can be more robust to real-world cases.
The first learning-based works applied classical methods such as Histogram of Oriented Gradients (HoG) and Support Vector Machines (SVM)~\cite{barnes2015exploiting,jensen2015traffic}. More recently, deep neural networks (DNNs) have been shown quite effective in various autonomous driving tasks, including traffic light detection. Some works show that generic object detectors, such as YOLO~\cite{redmon2016you} and Faster R-CNN~\cite{ren2015faster} are effective for traffic light detection~\cite{possatti2019traffic}. Behrendt \textit{et al.}~\cite{behrendt2017deep} modified the training procedure of YOLO~\cite{redmon2016you} to better handle issues more specific to traffic light detection, such as the small size of the objects of interest. In~\cite{pon2018hierarchical}, the authors modify the architecture and the mini-batch selection mechanism of Faster R-CNN~\cite{ren2015faster} to train it to detect traffic lights and signs simultaneously. Although detection methods based on deep learning are effective, they require large datasets, which are expensive to annotate. Moreover, they suffer from intrinsic data imbalance in traffic light detection datasets.

In this context, some works have been proposed to reduce the data imbalance and the effort required to build datasets. First, there are several tools~\cite{vott2018microsoft, scalabel2018berkeley} that attempt to mitigate the costs of annotating databases for detection tasks. In addition, many people are investigating (semi-)automatic techniques to aid the annotation process, some of them including human-in-the-loop~\cite{wang2018cvpr}.
Second, there are some works~\cite{oquab2015cvpr, sangineto2018pami} on weakly-supervised object detection that try to leverage the massive amounts of data annotated for classification to perform detection tasks. Moreover, few-shot learning~\cite{chen2018aaai, kang2018arxiv} has also been applied, to reduce the need for large amounts of collected data.
Lastly, data imbalance is a well-known issue and its impact on learning-based methods has been widely investigated even before deep learning~\cite{he2008tkde}. The standard tricks, that usually provide limited robustness, can be roughly categorized into two techniques: data re-sampling and cost-sensitive learning. In the deep learning context, some works~\cite{huang2016cvpr, wag2017neurips, zhou2018kdd} have investigated the effectiveness of these methods when dealing with imbalanced data as well as how to learn deep representations that take the imbalance into account. In~\cite{tabelini2019ijcnn}, the authors propose a method for training deep traffic sign detectors that does not require any target domain real images, where templates are overlaid on arbitrary natural images to generate training samples.
Nonetheless, there is still a need for further investigation and for solutions capable of handling these issues altogether, particularly when it comes to traffic light detection.
To fulfill this gap, we propose a method to generate synthetic data that does not require human-made annotations nor real data, unlike previous works, and can be used to train deep traffic light detectors with performance on par with models trained on datasets of real images. Our method is compared a state-of-the-art method~\cite{tabelini2019ijcnn}, which also does not require human-made annotations nor real data, and to the vanilla traffic light detection using deep learning \cite{ren2015faster} and real data from traffic scenes.

\section{Traffic light detection using synthetic context}

In order to avoid the arduous processes of acquiring and annotating real-world data, the proposed approach combines simple and non-realistic artificial 3D shapes and natural images. The artificial 3D shapes serve as foreground, describing the main elements of a traffic scene (e.g., traffic lights, lanes, poles, and vehicles), while the non-traffic-related natural images are used as background. The artificial foreground with traffic light annotations (i.e., the bounding box and the state) are generated automatically with simple and non-realistic computer graphics techniques. By combining a large variety of artificial foregrounds and natural backgrounds, a training dataset is built. This image collection is then used to train a deep detector to localize and classify traffic lights in real-world image samples. Figure~\ref{fig:overview}\footnote{This image presents modified images from COCO~\cite{lin2014microsoft} dataset which can be freely shared and modified under the Attribution License, available in \url{https://creativecommons.org/licenses/by/3.0/}. The figure, as others further presented in this work, also presents a Udacity's~\cite{gonzalez2018annotated} image free to be published under the MIT license.} summarizes the proposed method.

\subsection{Backgrounds}

The background image set comprises a large variety of natural scenes, like food on the table, animals in nature, people playing sports, and more. Therefore, an attractive option is to exploit large publicly available datasets, both because of their diversity (in case of a general object recognition/detection dataset) and their size. Moreover, given its use and to avoid the inclusion of false positives in the training data, we constrain this set to have only non-traffic-related images, filtering images that may fall into this category.

\subsection{Foregrounds}
\label{foregrounds}

Traffic context scenes can be simulated by modeling and combining a few basic traffic elements, such as roads, poles, vehicles, and traffic lights. The main idea is to reproduce a driver's view of a road intersection signalized by traffic lights in a simple manner. For the generation of the foreground image set, the open-source \textit{Processing} environment~\cite{processing} was used. \textit{Processing} is a software sketchbook oriented to program computer graphics applications, which enables combining different artificial elements by controlling a set of empirically/randomly selected parameters. The most relevant parameters control the position of the observing camera; the directions where the road intersection can lead to; the number of lanes of the roads; the presence or not of a crosswalk; the number, colors and positions of the vehicles in each lane; the road side where the traffic light poles are placed; the poles format; the number, the position, the angle and the state of each traffic light in pole; and, finally, the direction of light. Most traffic elements are modeled through simple geometric forms such that not much time is required to implement them. The components of a scene are described in the following paragraphs.

\begin{figure}
\begin{center}
   \includegraphics[width=0.65\columnwidth]{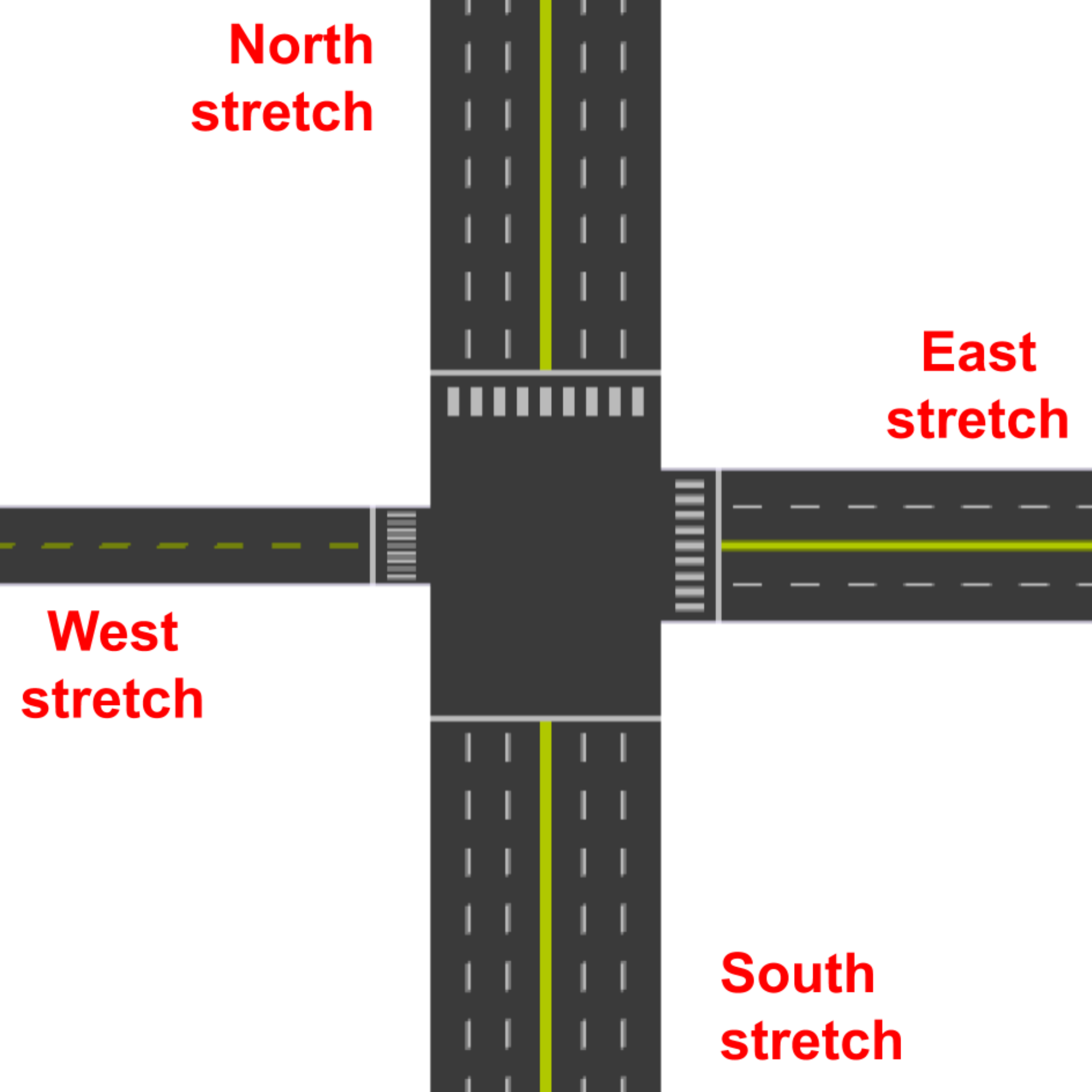}
\end{center}
   \caption{Illustration of a road composed by four stretches: south stretch, west stretch, north stretch, and east stretch.}
\label{fig:road}
\end{figure}

\paragraph{Road} A road is composed of stretches represented by a rectangle of $20H\times{lW}$ each, where $l$ corresponds to the number of lanes in the stretch and is uniformly drawn from $\{2, 4, 6\}$. Roads have two ways, each with $\frac{l}{2}$ lanes. Over the stretch, thin rectangles are also used to represent lane separators and crosswalks. Four different stretches can be generated. They can be referred as ``south'', ``west'', ``north'', and ``east'' stretches, as illustrated by Figure~\ref{fig:road}. Each scene presents a road with two to four stretches. The probabilities of generating each stretch are: 100\% for the south, 80\% for the north and west, and 100\% for the east if the previous two are not generated, otherwise 80\%. The west, north, and east stretches are generated similarly to the south stretch, but they are respectively rotated by 90, 180, or -90 degrees around their extremity so that they result in a road intersection with the configuration depicted by Figure~\ref{fig:road}.

\paragraph{Traffic light poles} Poles are represented by cylinders and are randomly positioned on the left or right side of the road stretch's extremes. Their axes are orthogonal to the road and may have a horizontal extension at their top. A pole's axis can have up to one traffic light, while its horizontal extension can have up to two.

\begin{figure}
\centering
\subfloat[]{\includegraphics[width=0.14\textwidth]{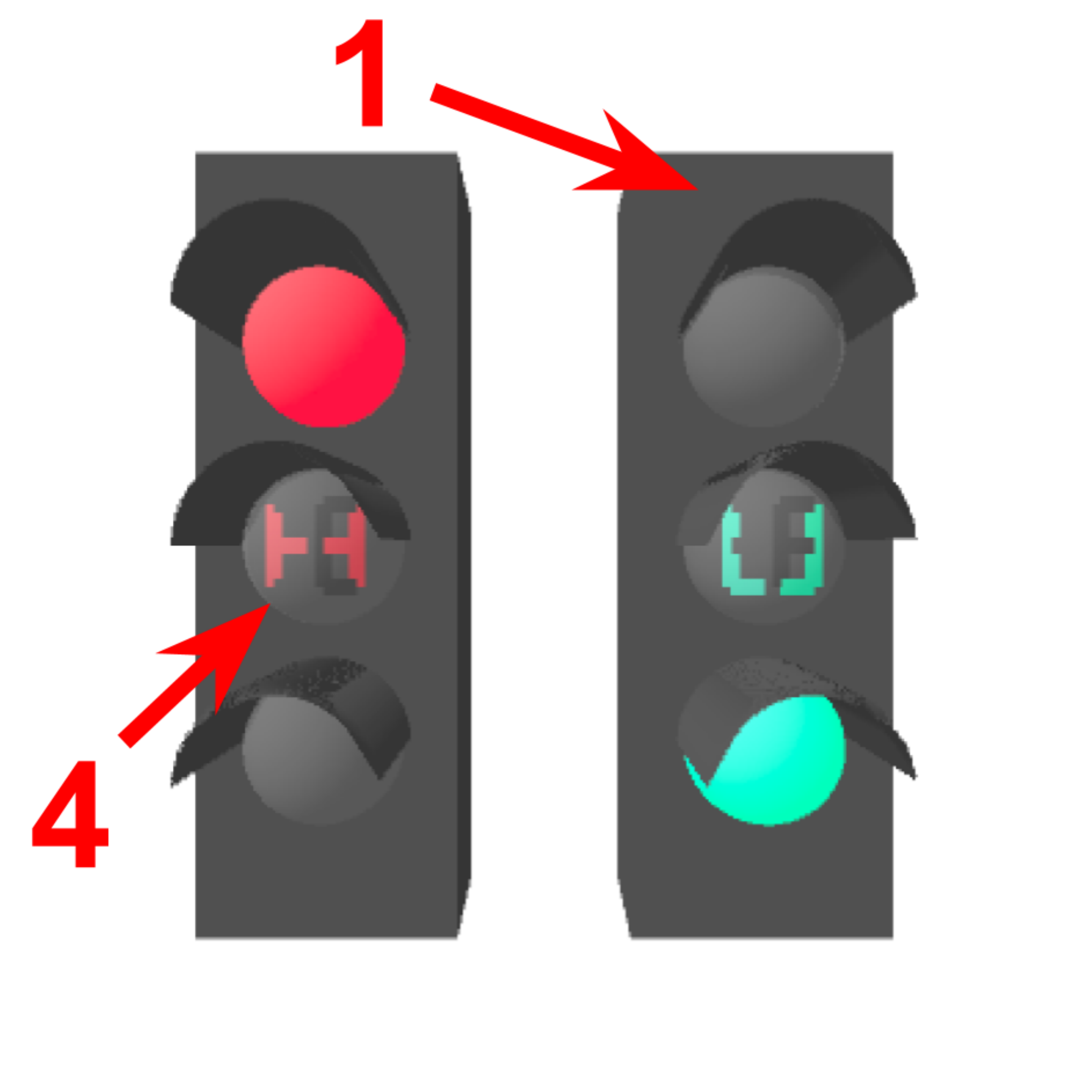}%
\label{fig:timer}}
\subfloat[]{\includegraphics[width=0.14\textwidth]{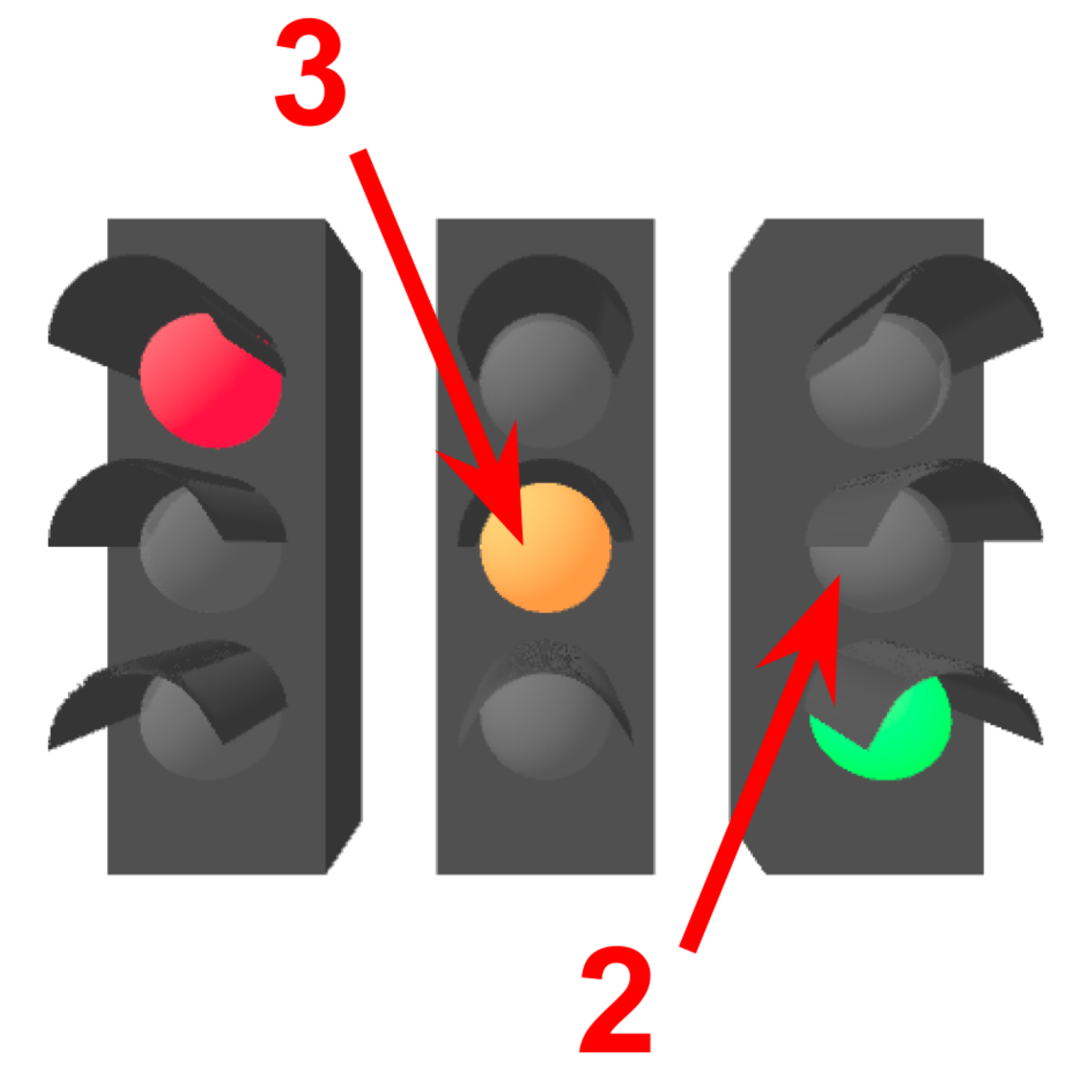}%
\label{fig:lighted}}
\subfloat[]{\includegraphics[width=0.14\textwidth]{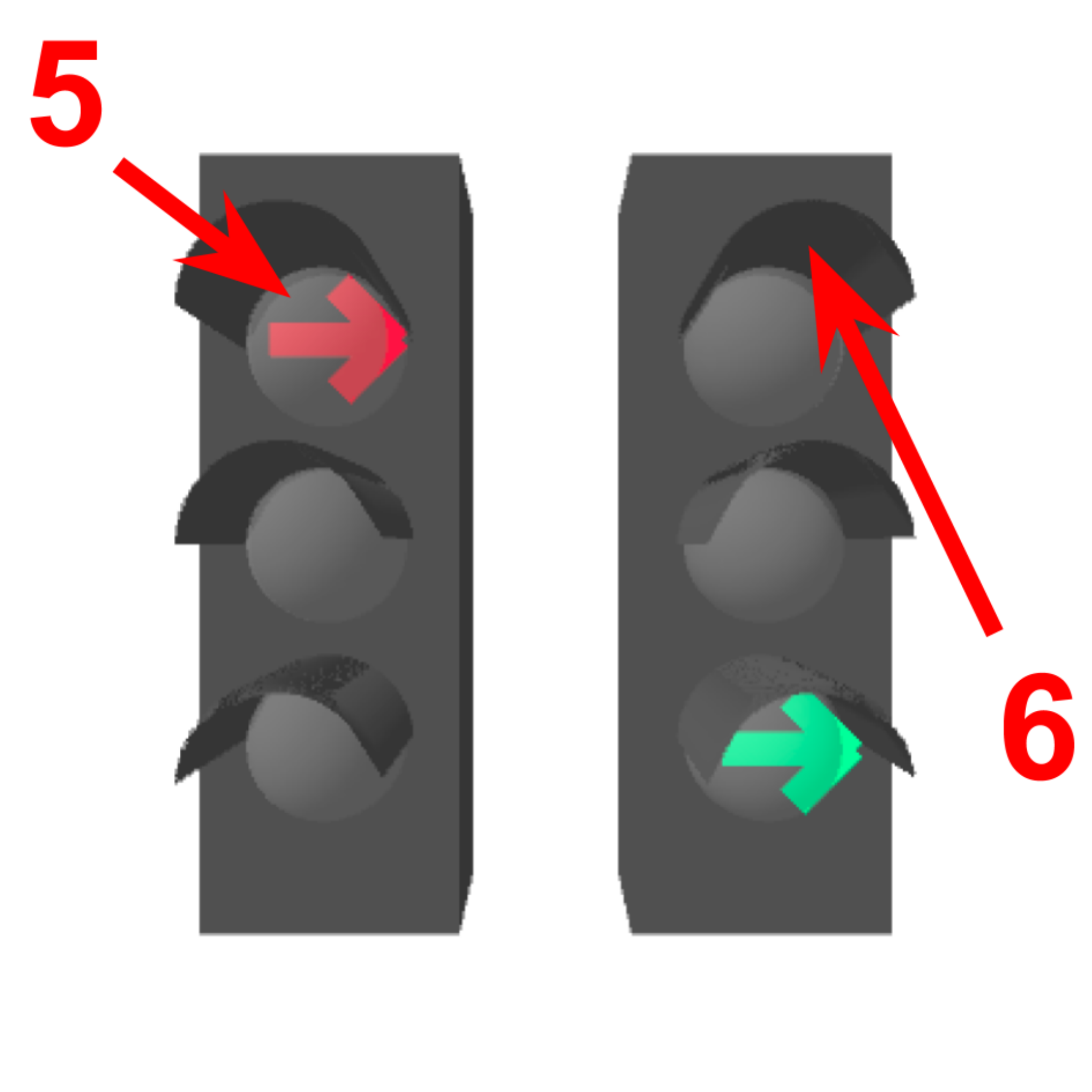}%
\label{fig:arrow}}
\caption{Examples of modeled traffic lights: (a)~traffic lights containing a timer; (b)~traffic lights with fully-lighted bulbs; (c)~directional traffic lights. Represented components: 1.~Traffic light's body; 2.~Off-state bulb; 3.~Fully-lighted bulb; 4.~Bulb with timer; 5.~Bulb with arrow; 6.~Bulb covering.}
\label{fig:models3d}
\end{figure}

\paragraph{Traffic lights} The design of the modeled traffic lights, exemplified in Figure~\ref{fig:models3d}, corresponds to the three-bulb vertical black model, and each traffic light is assigned with a red, yellow, or green state with equal probability in order to avoid the traditional imbalance over the yellow state. A traffic light model is composed of a black box intercepted by three spheres representing the bulbs. The following models are generated: one fully-lighted bulb and a lighted timer in the central bulb (Figure~\ref{fig:timer}); one fully-lighted bulb and two off bulbs (Figure~\ref{fig:lighted}); and one bulb with only a lighted arrow (Figure~\ref{fig:arrow}). The figure shows each of the following components of the 3D models: 1.~\textit{Traffic light's body}: black box; 2.~\textit{Off-state bulb}: dark sphere; 3.~\textit{Fully-lighted bulb}: sphere simulating an emitting colored material in random tone of the color of the traffic light state; 4.~\textit{Bulb containing timer}: a timer is represented by two digits, composed by five dark-colored segments each. The segments are lighted randomly, therefore the final figure might not be a real digit. The segments are modeled by boxes (with emitting material if lighted) and the bulb has intensified transparency so that the timer inside it can be visualized; 5.~\textit{Bulb containing arrow}: similar to the previous one, but with lighted segments picturing an arrow directed to the left or right; 6.~\textit{Bulb covering}: transversal segments of cylinder covering the bulbs.

\paragraph{Vehicles} Vehicles are introduced into the scene by loading (from a publicly available object file\footnote{\url{https://free3d.com/3d-model/bmw-x5-1542.html}.}) a 3D car model without texture and placing several instances into the sketch. Each instance is positioned on the road, between the lane separators, and have their colors randomly set so that the generated scene comprises vehicles with different colors. Back headlights are also modeled through placing a yellow box with emitting material inside a red translucent box and, then, added to each instance so that the deep detector could distinguish them from traffic lights. Dark ellipses under vehicles simulate their shadow over the pavement.

\paragraph{Light} A directional light is added in random directions varying along and laterally to the road, but always coming from above the road. Therefore, each scene has its own illumination configuration.

\paragraph{Camera} The camera position is defined to simulate the view of a driver in the scene. The camera is placed in the center of one of the possible three south road stretch right lanes ($0.5W$, $1.5W$ or $2.5W$, since the lane width is $W$). The position along the road is set to be within a certain distance ($8.9H$-$21H$) from the road intersection where the traffic lights are. This approach generates traffic lights viewed from different distances throughout scenes. The position over the road is set to represent the height of a driver's eye ($200$-$300$ pixels distant from the road surface). The camera points toward the center of the road intersection. \\

When each traffic light associated with the south stretch is placed on a position of the tridimensional scene, the bidimensional positions of its frontal face's extremities in the final image are calculated and defined as its bounding box coordinates. These traffic lights are also labeled according to their states. Two procedures are considered to avoid traffic lights with high level of occlusion to be labeled: (I) if 50\% or more of the bounding box area is out of the image limits, the traffic light is not labeled; (II) a tridimensional bounding box is considered for the vehicles in the south stretch, then the bidimensional position of its vertexes in the image is calculated. The obtained four extreme positions define the vehicle's 2D bounding box. If a vehicle's bounding box overlaps 50\% or more of the area of a traffic light's bounding box, the vehicle is not placed on the scene. 

Once complete, the scene is saved as an image, such as shown in Figure~\ref{fig:overview} (``Artificial 3D traffic contexts''). The background of the synthetic image (region without information) is set as transparent so that it shows the non-traffic-related natural image as background after blending.

\subsection{Data generation}
\label{sec:data_generation}

The assembling of a training image starts by randomly selecting a foreground image to be blended into a random background image. Once background and foreground are selected, they undergo an image augmentation process in order to increase data variability within the dataset. Basically, the augmentation comprises four steps: background and foreground individual brightness transformations, foreground histogram noising, blending and blurring. All augmentation parameters were selected empirically to avoid degenerated images. \added{Both foreground and background images are rescaled to the final dimensions of $1280\times960$ pixels to aid the visual inspection of the images when choosing the range of the augmentation parameters.}

The brightness transformation involves adding and multiplying random values to the original images' pixel values. For the background image, a real value sampled from the interval $[-120, 120)$ is added to each channel of all pixels. The resulting values are then multiplied by a coefficient within $[0.75, 1.25)$. Consider $A_B$ and $A_F$ as the values added to the pixels of the selected background and foreground, respectively. The same sequence of operations is applied to both, background and foreground images, with the exception that $A_F = A_B + 40$, so that the foreground tends to be slightly highlighted over the background.

Since the foreground is artificially generated by a simple computer graphics process, the color of each shape is originally uniform along all their extension. To make the scene more realistic and increase data variability, a histogram noising process is performed on the selected foreground. A random integer value within $[-15, 15)$ is added in a pixel-wise fashion to each channel of the image, therefore adjacent pixels have different color intensities, even when composing the same shape.

Besides color uniformity, the artificial shapes have sharp edges, i.e., there is no smooth transition between the scene elements, such as the crosswalk and the road pavement. To address this, a Gaussian blur filtering is applied to the foreground, in which the standard deviation of the filter is randomly drawn from the range $[0, 3)$.

Smooth transition is also desirable between foreground's limits and the background. Therefore, the blending procedure was performed considering a mask with smooth transitions in the borders of the objects. The smooth transition was generated by applying a sequence of erosion operations in which each of them introduces a different level of opacity. More specifically, let $M_F$ denote the original foreground's mask (encoded as a 0-1 float image) and $M$ a new foreground's mask to be generated. First, $M$ is defined as $M = \frac{M_F}{3}$. Then, $M_F$ is eroded twice by a $3\times3$ square kernel and $\frac{1}{3}$ of each result is added to $M$. This makes $M$ a mask in which the object's border pixels are less intense. The blending procedure produces an image $I$ which results from overlapping the background $B$ with the foreground $F$ in correspondence to this new mask, i.e., $I = (1 - M)B + MF$.

Finally, the blended image is submitted to another blur filtering in order to increase the smoothness of the final arrangement and also the data variability. The standard deviation is once more randomly selected from the range $[0, 3)$.

\section{Experimental methodology}
\label{experimental_methodology}

The experimental evaluation aims to assess the utility of a synthetic dataset generated by the proposed approach. In this context, the performance of a deep detector trained only on the synthetic-generated data is measured on several well-known datasets, and compared with reasonable baselines as well as with a deep detector trained with real-world traffic scenes, which can be viewed as an empirical upper-bound for cross-database experiments. The following subsections describe the training, validation and test datasets used for the experiments, the base datasets used to assemble them, the metrics for performance evaluation, the experimental setup, the conducted experiments and the computational resources used to run them. \added{The concept of \textit{box-validation} (used to provide fair comparisons among different annotation schemes) is also introduced. Table~\ref{table:datasets} summarizes information about the used datasets that are described in the next subsections. The foreground images, the source code for the generation of the datasets and the trained models are publicly available\footnote{\camreadyadd{\url{https://github.com/Jpvmello/traffic-light-detection-synthetic-context}.}}.}

\begin{table*}[ht]
  \centering
  \caption{\added{Details of the datasets (synthetic, real-world, and hybrid) with their respective training, validation, box-validation, and test sets.}}
  \label{table:datasets}
  \resizebox{\textwidth}{!}{%
    \begin{tabular}{lll|ccc|ccc}
      \toprule
      \multirow{2}{*}{\textbf{Dataset}} & \multirow{2}{*}{\textbf{Base Dataset}} & \multirow{2}{*}{\textbf{Set}} & \multicolumn{3}{c}{\textbf{Images}} & \multicolumn{3}{|c}{\textbf{Traffic lights}} \\
      & & & \textbf{Size} & \textbf{Positives} & \textbf{Negatives} & \textbf{Red} & \textbf{Green} & \textbf{Yellow} \\ \midrule
      \multirow{2}{*}{Fully Contextualized} & \multirow{2}{*}{COCO + context} & Training & \multirow{2}{*}{1280$\times$960} & 70,000 & 0 & 46,432 & 46,619 & 47,064 \\
      & & Validation & & 7,000 & 0 & 4,701 & 4,702 & 4,543 \\
      \multirow{2}{*}{Uncontextualized} & \multirow{2}{*}{COCO + 3D models} & Training & \multirow{2}{*}{1280$\times$960} & 70,000 & 0 & 46,432 & 46,619 & 47,064 \\
      & & Validation & & 7,000 & 0 & 4,701 & 4,702 & 4,543 \\
      \multirow{2}{*}{Templates Only} & \multirow{2}{*}{COCO + 2D templates} & Training & \multirow{2}{*}{1280$\times$960} & 70,000 & 0 & 46,432 & 46,619 & 47,064 \\
      & & Validation & & 7,000 & 0 & 4,701 & 4,702 & 4,543 \\
      \multirow{2}{*}{Positive Backgrounds} & \multirow{2}{*}{BDD100K positives + 3D models} & Training & \multirow{2}{*}{1280$\times$960} & 70,000 & 0 & 46,432 & 46,619 & 47,064 \\
      & & Validation & & 7,000 & 0 & 4,701 & 4,702 & 4,543 \\
      \multirow{2}{*}{Negative Backgrounds} & \multirow{2}{*}{BDD100K negatives + 3D models} & Training & \multirow{2}{*}{1280$\times$960} & 70,000 & 0 & 46,432 & 46,619 & 47,064 \\
      & & Validation & & 7,000 & 0 & 4,701 & 4,702 & 4,543 \\
      \multirow{2}{*}{Real-world Reference} & \multirow{2}{*}{DTLD} & Training & \multirow{2}{*}{1280$\times$960} & 70,000 & 0 & 46,432 & 46,619 & 47,064 \\
      & & Validation & & 7,000 & 0 & 4,701 & 4,702 & 4,543 \\ \midrule
      \multirow{2}{*}{LISA\_train+test} & \multirow{2}{*}{LISA} & Box-validation & \multirow{2}{*}{640$\times$480} & 1,277 & 125 & 2,199 & 1,509 & 100 \\
      & & Test & & 18,971 & 4,615 & 29,731 & 20,889 & 1,402 \\
      \multirow{2}{*}{LISA\_test} & \multirow{2}{*}{LISA} & \multirow{2}{*}{Test} & \multirow{2}{*}{640$\times$480} & \multirow{2}{*}{7,473} & \multirow{2}{*}{3,481} & \multirow{2}{*}{9,846} & \multirow{2}{*}{7,717} & \multirow{2}{*}{457} \\
      & & & & & & & & \\
      \multirow{2}{*}{Udacity-} & \multirow{2}{*}{Udacity} & Box-validation & \multirow{2}{*}{1920$\times$1200} & 447 & 1,052 & 666 & 463 & 29 \\
      & & Test & & 4,027 & 9,474 & 6,176 & 3,777 & 199 \\
      \multirow{2}{*}{Udacity+} & \multirow{2}{*}{Udacity} & Box-validation & \multirow{2}{*}{1920$\times$1200} & 447 & 1,052 & 663 & 463 & 28 \\
      & & Test & & 4,027 & 9,474 & 6,162 & 3,811 & 178 \\
      \multirow{2}{*}{LaRA} & \multirow{2}{*}{LaRA} & Box-validation & \multirow{2}{*}{640$\times$480} & 593 & 524 & 512 & 361 & 6 \\
      & & Test & & 5,339 & 4,723 & 4,768 & 3,020 & 52 \\
      \multirow{2}{*}{Proprietary} & \multirow{2}{*}{Proprietary} & Box-validation & \multirow{2}{*}{1280$\times$960} & 269 & 130 & 191 & 296 & 19 \\
      & & Test & & 2,426 & 1,172 & 1,622 & 2,627 & 247 \\ \bottomrule
    \end{tabular}
  }
\end{table*}

\subsection{Backgrounds datasets}
\label{sec:natural_backgrounds_dataset}

This subsection describes the Microsoft Common Objects in Context (COCO) collection~\cite{lin2014microsoft}, from which the background images are selected, and the Berkeley DeepDrive (BDD100K), a dataset with real-world traffic scenes used as backgrounds for comparison with the use of non-traffic-related backgrounds.

\subsubsection{Microsoft Common Objects in Context (COCO)}

The training partition of the 2017 version of COCO is used as a source for non-traffic-related natural images to the proposed method. The dataset comprises a total of 328k images with 91 labeled categories of common objects. For the experiments, the dataset was filtered not to contain the following traffic elements: ``traffic light'', ``bicycle'', ``car'', ``bus'', ``motorcycle'', ``truck'', and ``stop sign''. From the filtered set, a subset of 37k randomly-selected images with the smallest dimension equal or higher to 120 pixels composes the natural data used as background for the experiments (30k designated for training and 7k for validation). Images were rescaled to have at least 480 height and 640 width, without altering their aspect ratios. Subsequently, the central pixels were cropped so that the images have dimensions of $640\times480$ pixels. Since COCO is available online, it is assumed that any person using the proposed method, or a variation of it, could have access to the data, including its filter tags.

\subsubsection{Berkeley DeepDrive (BDD100K)}

The BDD100K dataset~\cite{bdd100kberkeley,yu2020bdd100k} consists of 100k videos recording 40 seconds of more than 50k driving rides conducted in different cities from the USA. The dataset is prepared to be used for ten different tasks, incluing detection, segmentation and tracking of elements in the traffic domain. For image tasks, the interest of this work, the frame corresponding to the 10\textsuperscript{th} second of each video is annotated, resulting in an image dataset with 100k images with dimensions of $1280\times720$ pixels.

For this work, only images from daytime and dawn/dusk scenes were considered. They were rescaled to be 960 pixels high and had their central $1280\times960$ pixels cropped. The resulting set was split into one comprising only the positive images (with at least one labeled traffic light), and other containing only the negative images (with no labeled traffic lights). The resulting set of positive backgrounds have 23,850 images, while its counterpart contains 23,941 images. Random 7k images from each of these sets were designated for validation and the remaining ones for training.

\subsection{Training and validation datasets}

This subsection describes the training and validation datasets assembled for the experiments. They include (i) synthetic-generated datasets, denoted as Fully Contextualized (Ours), Uncontextualized and Templates Only, (ii) a real-world traffic dataset, an adaptation of the DriveU Traffic Light Dataset (DTLD), used to train a strong baseline denoted as Real-world Reference, and (iii) hybrid datasets denoted as Positive Backgrounds and Negative Backgrounds, combining synthetic foregrounds with traffic-related backgrounds from the BDD100K.

\subsubsection{Fully Contextualized (Ours)}
\label{sec:fully_contextualized}

The Fully Contextualized dataset is also denoted as Ours since it refers to the dataset generated through the proposed method. COCO backgrounds and samples from a set of 20k foregrounds generated with dimensions of $640\times480$ pixels were randomly combined to assemble the training set, according to the data generation process described in Section~\ref{sec:data_generation}. This process resulted in a set with the arbitrary number of 70k images with dimensions of $1280\times960$ pixels and 46,432 traffic lights labeled as red, 46,619 as green and 47,064 as yellow.

The corresponding validation set, containing 7k images, was assembled in a similar manner. However, no repeated backgrounds and foregrounds occur, i.e., each of the 7k validation backgrounds is combined with a unique sample from a set of 7k generated foregrounds (different from the ones designated for the training set). This validation set contains 4,701 traffic lights labeled as red, 4,702 as green and 4,543 as yellow.

\subsubsection{Uncontextualized}
\label{sec:uncontextualized}

\begin{figure}
\centering
\subfloat[Fully Contextualized]{\includegraphics[width=0.48\columnwidth]{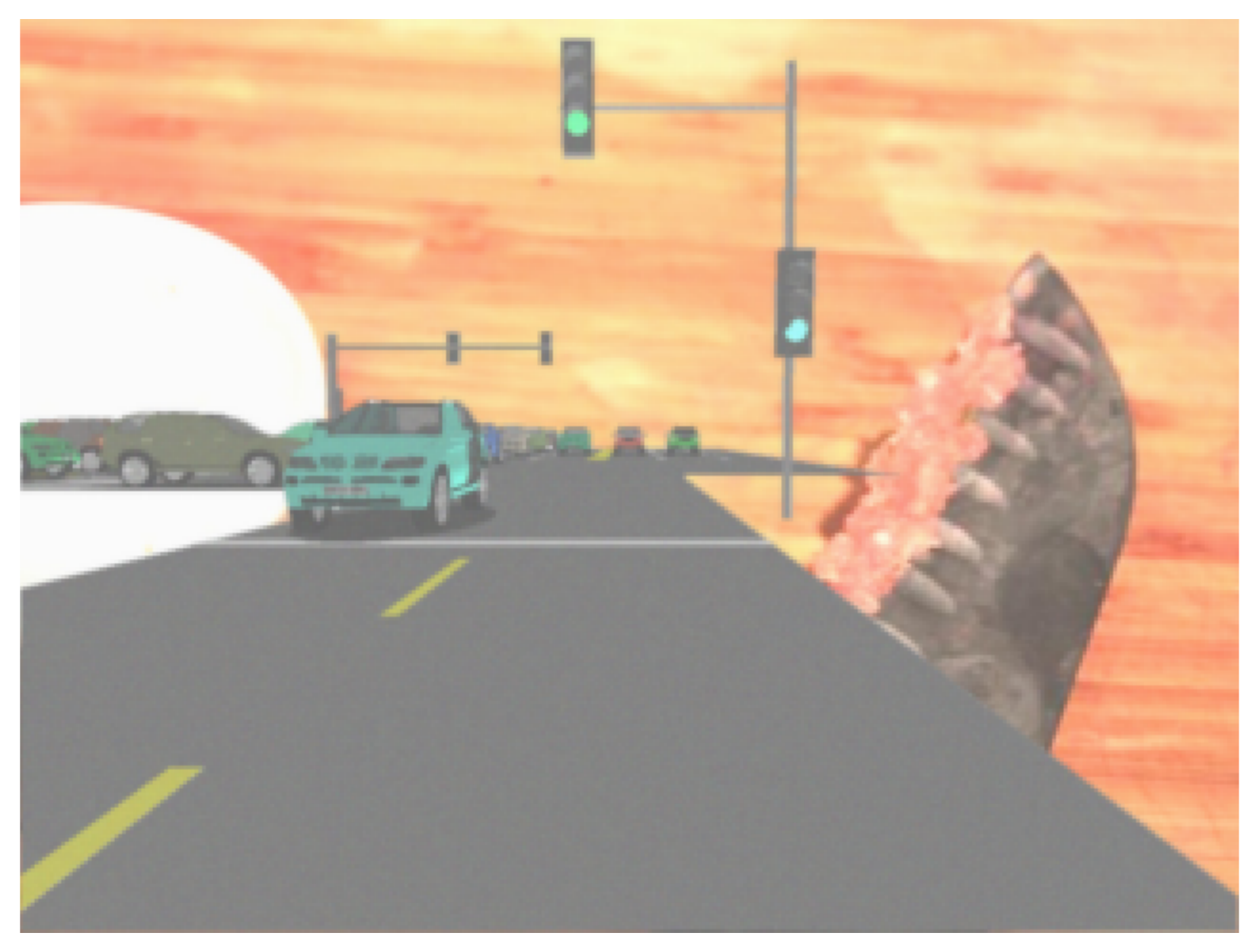}%
\label{fig:with_context}}  
\subfloat[Uncontextualized]{\includegraphics[width=0.48\columnwidth]{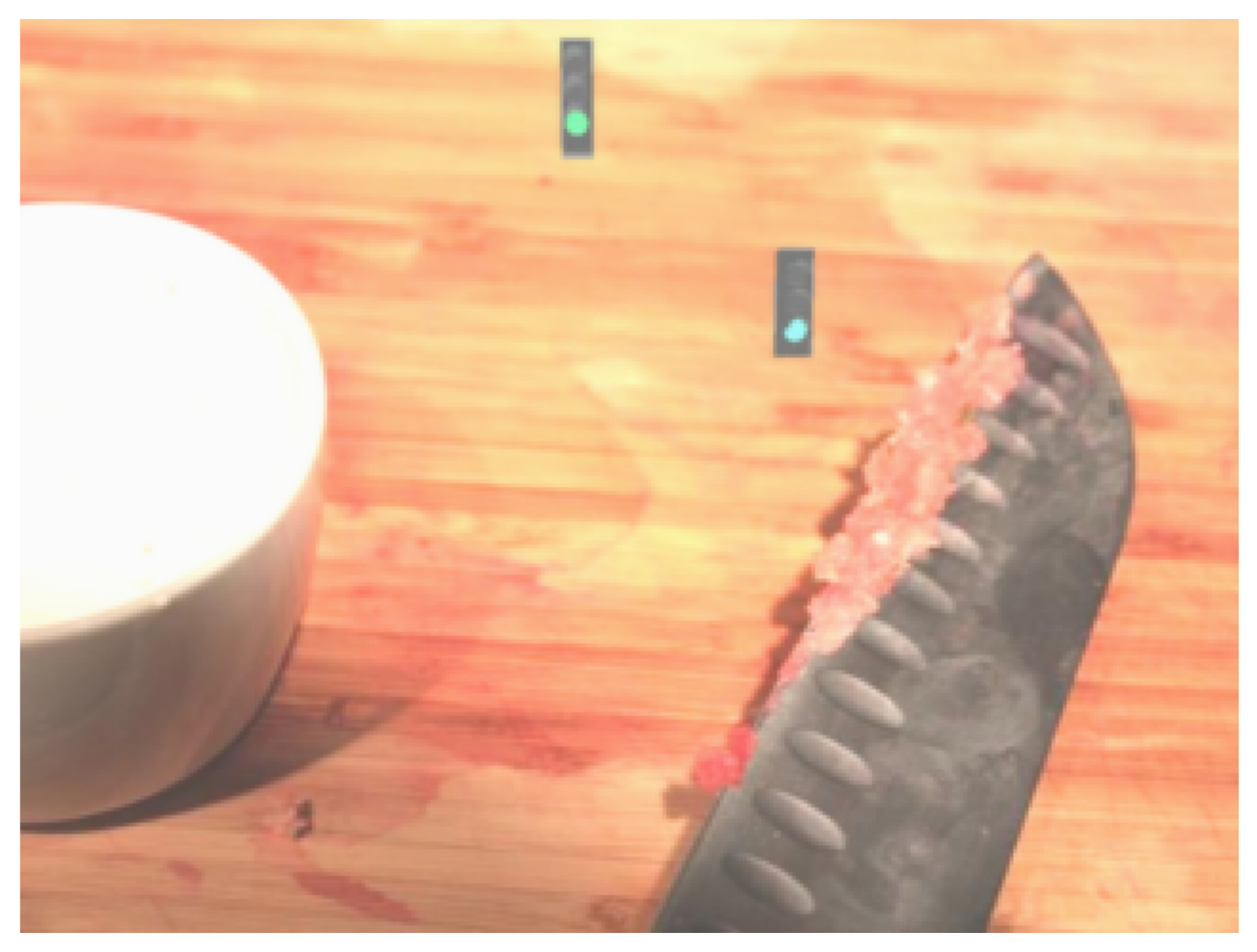}%
\label{fig:with_outcontext}}
\caption{Samples from the Fully Contextualized and Uncontextualized datasets.}
\label{fig:context_levels}
\end{figure}

This dataset (both training and validation sets) has the exact same traffic scenes of the Fully Contextualized dataset (Section~\ref{sec:fully_contextualized}), but drawing only the traffic lights instead of the whole traffic context. Some samples are shown in Figure~\ref{fig:context_levels}.

\subsubsection{Templates Only}

\begin{figure}
\centering
\subfloat[]{\includegraphics[width=0.14\textwidth]{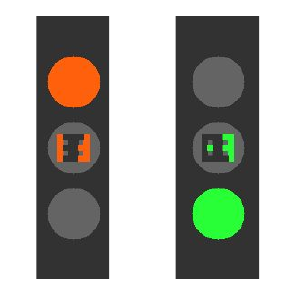}%
\label{fig:timer2d}}
\subfloat[]{\includegraphics[width=0.14\textwidth]{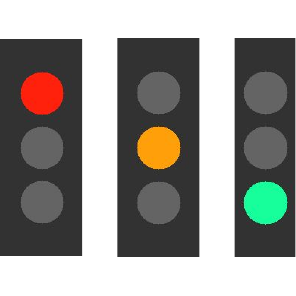}%
\label{fig:lighted2d}}
\subfloat[]{\includegraphics[width=0.14\textwidth]{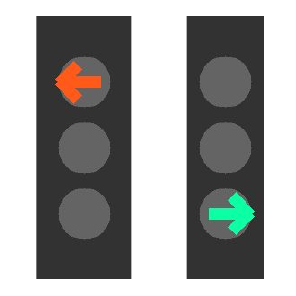}%
\label{fig:arrow2d}}
\caption{Examples of modeled templates: (a)~traffic lights containing a timer; (b)~traffic lights with fully-lighted bulbs; (c)~directional traffic lights.}
\label{fig:templates2d}
\end{figure}

The Templates Only dataset is paired with the Uncontextualized dataset in both training and validation sets (presented in Section~\ref{sec:uncontextualized}) in terms of labeling (classes and dimensions) and augmentation, but it replaces the 3D traffic light models by randomly generated 2D templates with fully-lighted bulb (Figure~\ref{fig:lighted2d}), with a timer (Figure~\ref{fig:timer2d}), or with a directional arrow (Figure~\ref{fig:arrow2d}). They were designed to look as similar as possible to the faces of the 3D models represented in Figure~\ref{fig:models3d}, reproducing the ranges of possible width and height, diameter of the bulb, possible color tones, among others.

Before blending, each template's perspective is transformed so that the sides farthest from the center of the final image are smaller than the respective opposite sides. Then, it is rotated by an angle in degrees uniformly drawn from $[-3.6, 3.6]$, resized to fit the respective paired label and placed in a transparent image to be blended to a background image, maintaining the paired label's coordinates and dimensions.

\subsubsection{\added{Positive Backgrounds and Negative Backgrounds}}

\added{The Positive Backgrounds dataset is also paired with the Uncontextualized dataset, but replacing the COCO non-traffic-related backgrounds by the positive traffic-related backgrounds from BDD100K. The Negative Backgrounds dataset is equivalently generated, except by the fact that it uses the negative BDD100K backgrounds set.}

\subsubsection{Real-world Reference: DriveU Traffic Light Dataset (DTLD)}

DTLD~\cite{fregin2018driveu} contains 43,132 images with dimensions of $2048\times1024$ pixels and more than 230k annotations of traffic lights. Its images are part of recordings produced in 11 German cities. The annotations provide information about the following features of the traffic lights: (i) face orientation, (ii) occlusion and relevancy to the route of the vehicle which was used to record the dataset, (iii) axial orientation, (iv) number of light bulbs, (v) state and (vi) pictogram (fully-lighted, arrow, pedestrian, etc.).

The dataset's images were processed and filtered before use. First, the 960 superior rows and the 1280 central columns of each image were cropped. Then, the cropped area was rescaled to half its dimensions, resulting in an image with dimensions of $640\times480$ pixels without distortion. Next, the dataset was filtered to select only images containing traffic lights with features of interest, namely: (i) frontal face view, (ii) all levels of relevance, (iii) vertical, (iv) three bulbs, (v) red, yellow and green, and (vi) with fully-lighted or arrow bulbs. The filtered dataset has 33,374 images and a total of 25,956, 47,919 and 2,572 traffic lights in the red, green and yellow states, respectively.

To compensate class imbalance, the dataset was reorganized to replicate images from the least frequent classes as described next. Let $Y$, $R$ and $G$ be subsets of images containing, respectively: at least one yellow traffic light; at least one red and no yellow traffic lights (note that green may also occur); and at least one green and no yellow traffic lights (red also possible). Balance was achieved by selecting one image from each set repeated times and assigning them to a new set. This was repeated until a total of 70k images were selected. This results in a final training dataset with 46,496 annotated yellow traffic lights, 56,521 red and 58,993 green. For the experiments, the same augmentations applied to the COCO backgrounds, described in Section~\ref{sec:data_generation}, were also applied to this dataset.

\added{A validation set based on DTLD was not assembled. Instead, a best-case scenario was considered for each test dataset. For this, the Real-world Reference was validated on each box-validation set, which comprises a small portion of its respective test dataset, as further described in Section~\ref{sec:test_and_boxval_datasets}. Therefore, its validation was conducted with a dataset-dependent positive bias so that the proposed method performance can be compared with the best performance that the Real-world Reference can provide on each test set.}

\subsection{Test and box-validation datasets}
\label{sec:test_and_boxval_datasets}

\begin{figure}
\begin{center}
   \includegraphics[width=\columnwidth]{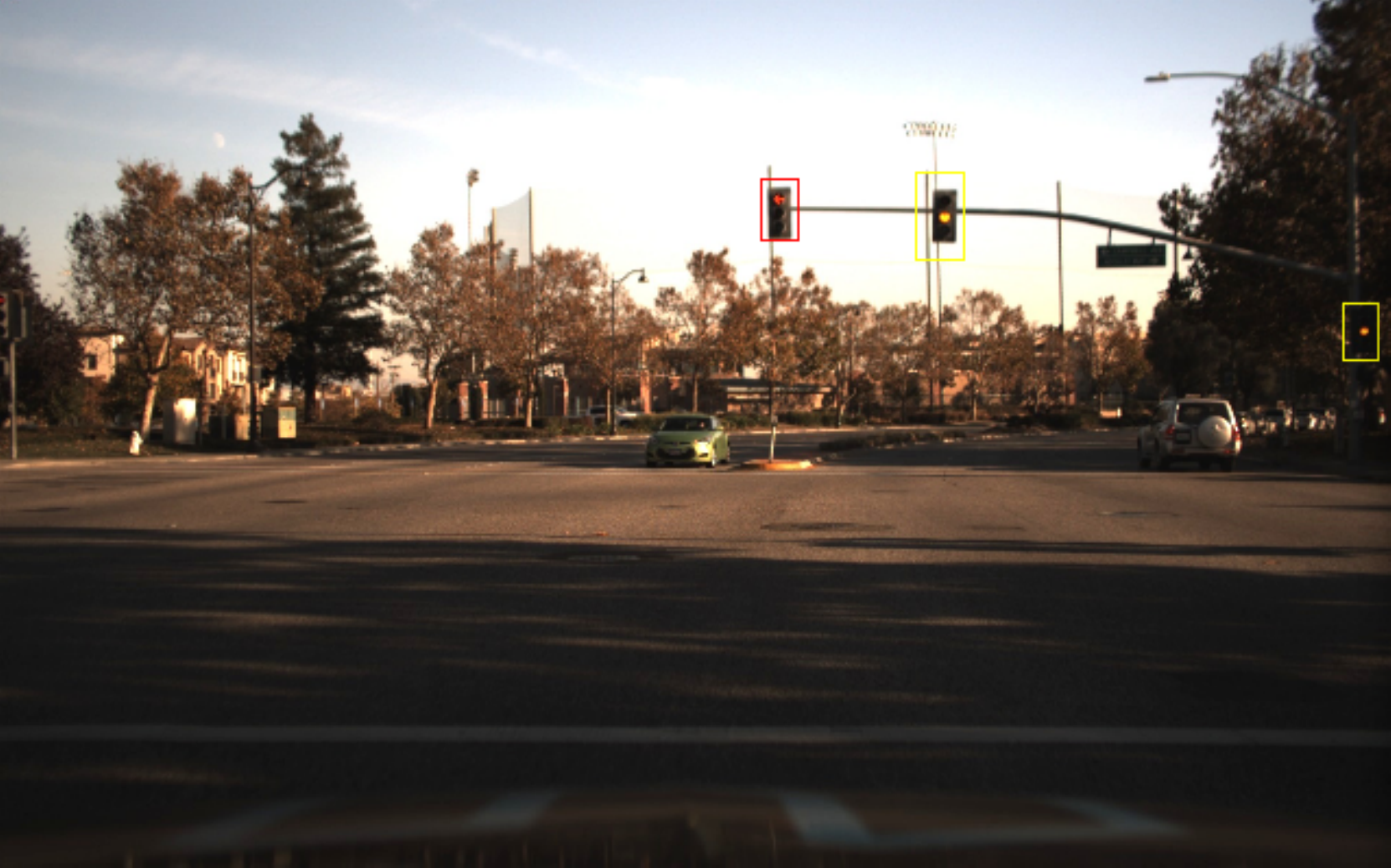}
\end{center}
   \caption{Sample from the Udacity dataset with the original ground truth bounding boxes. Note that their area considerably exceeds the actual traffic lights' areas.}
\label{fig:annotations}
\end{figure}

To evaluate the performance of the trained models, four datasets of real traffic images (LISA, Udacity, LaRA, and a proprietary one) were used, as described in the following subsections. \added{Each of the mentioned datasets is split into two subsets: a minority set with randomly selected 10\% of the dataset's positive and negative images (referred to as \emph{box-validation} set), and the majority set comprising the rest of the images (referred to as \emph{test set}. In an overview, the box-validation dataset -- as a sample of the test set -- serves to guide the adjustment of the predicted bounding boxes in order to compensate inaccurate annotations (as those in Figure~\ref{fig:annotations}) in the respective real-world test set. The use of the box-validation set is better explained in Section~\ref{sec:eval}}.

\subsubsection{LISA\_train+test and LISA\_test: Laboratory for Intelligent \& Safe Automobiles (LISA)}

The LISA traffic light dataset ~\cite{jensen2018lisa,jensen2016vision,philipsen2015traffic} contains day- and night-time traffic-related images with dimensions of $1280\times960$ pixels collected in San Diego, California, USA. The dataset is originally divided into train and test sets. The train set is subdivided into 13 day clips and 5 night clips, while the test set in 2 day sequences and 2 night sequences. Their images were rescaled to half of their original dimensions, i.e., $640\times480$ pixels. For this work, both the original daytime train and test sets were used to compose the box-validation and test sets. Traffic lights labeled as ``stop'' or ``stopLeft'' were considered as red. Those annotated as ``go'' or ``goLeft'' were redefined as green. Finally, those annotated as ``warning'' or ``warningLeft'' were taken as yellow. From the 14,034 images that compose the used original training day clips, 12,775 are positive. A total of 37,809 traffic lights are considered, from which 22,084 are red, 14,681 are green and 1,045 are yellow. In turn, the original test day sequences contain 10,954 images, from which 7,473 are positive, and present 9,846 red traffic lights, 7,717 green and 457 yellow. For the experiments, two test sets are considered: one corresponding to LISA's original test set (denoted as LISA\_test) and another one composed by both the original train (excluding the images used for validation) and test sets (denoted as LISA\_train+test).

\subsubsection{Udacity- and Udacity+}

The Udacity's public repository~\cite{gonzalez2018annotated} provides two datasets with several types of annotation. However, only the second dataset was annotated for traffic lights. Therefore, only the images of the second dataset are used for evaluation. This dataset has 15k images with dimensions of $1920\times1200$ pixels, referring to daytime traffic scenes from Mountain View, California.

Traffic lights labeled as occluded were not considered, resulting in 4,474 positive images and a total of 8,232 labeled red traffic lights, 5,639 greens and 278 yellows. However, almost 3k of the traffic lights were originally labeled more than once, causing overlapping bounding boxes. Thus, two distinct annotation patterns were considered to decide on overlapping bounding boxes: the first one, denoted as Udacity-, considers only the boxes with the smallest area, while the second one, denoted as Udacity+, considers the boxes with the biggest area. There are small differences on the number of traffic lights labeled as each state between each of the two patterns, due to the occurrence of overlapping boxes from more than one traffic light, but approximately 6,8k are labeled as red, 4,2k as green and little more than 200 as yellow.

\subsubsection{\textit{La Route Automatis{\'e}e} (LaRA)}
The LaRA traffic lights dataset~\cite{charette2013traffic} comprises 11,179 images with dimensions of $640\times480$ pixels from a video acquired through the traffic of Paris, France. Approximately 55\% of its images (5,932) are positive. It has a total of 5,280 annotations of red traffic lights (as ``stop''), 3,381 of greens (as ``go'') and only 58 in the yellow state (as ``warning'').

\subsubsection{Proprietary dataset}
A proprietary dataset \camreadyadd{(Intelligent Autonomous Robotic Automobile (IARA)\footnote{\camreadyadd{Available in \url{https://drive.google.com/drive/folders/1iATG5suB9bHnFi9x6XaWtjG-uzwsJ8kb}.}})} was also used for this work. It is composed of 3,997 traffic-related images of dimensions of $1280\times960$ pixels captured by a camera. This image set comprises 2,695 positive images. The dataset has a total of 1,813, 2,923 and 266 traffic lights in the red, green and yellow states, respectively.

\subsection{Experimental setup}

The models were trained using a publicly available Tensorflow implementation\footnote{\url{https://github.com/endernewton/tf-faster-rcnn}.} of a consolidated state-of-the-art object detector, Faster R-CNN~\cite{ren2015faster}, given the satisfactory results obtained by Torres \textit{et al.}~\cite{tabelini2019ijcnn}, using the also state-of-the-art ResNet-101~\cite{he2016deep} feature extractor. Faster R-CNN comprises a Convolutional Neural Network that proposes regions of interest as candidates of possible objects and two fully-connected networks, one for the bounding eu box regression and another for classification.

The anchor boxes scale and ratio sets were empirically defined as $\{2, 4, 8, 16, 32\}$ and $\{0.5, 1, 2\}$ respectively. The minimum overlap threshold of regions of interest is set to zero. Each model is trained for 70k iterations, so that each image is iterated once, with a batch size of 1. \added{This number of iterations was also empirically verified to be enough for convergence.} The learning rate is defined as $10^{-3}$ during the first 50k iterations and then it is decreased to 10\% of its original value for the rest of the training.

To increase the range of the traffic lights sizes during the training stage, the set of training image scales was defined as $\{480, 960\}$, i.e., each image is resized during training so that its smallest dimension becomes equal to one of those two values (randomly sampled). As the training samples have final dimensions of $1280\times960$, each image either keeps its original dimensions or is rescaled to half its dimensions. In turn, the test images were scaled during evaluation so that their smallest dimension is equal to 960, without change in their aspect ratios. In other words, images from the LISA, LaRA and the proprietary datasets were rescaled to $1280\times960$ pixels, while Udacity images were rescaled to $1536\times960$ pixels.

\subsection{Evaluation metrics and procedures}
\label{sec:eval}

The metrics adopted for the evaluation of the models were F1-score and mean Average Precision (mAP), derived from the precision, i.e., the ratio of correct predictions from all predicted bounding boxes, and recall, i.e., the ratio of correct predictions from all ground truths. To be considered correct, a prediction box must have an Intersection-over-Union (IoU) equal to or greater than 0.5 with a ground truth box, along with the correct classification. The precision and recall themselves may also be assessed for helping some results analysis.

The F1-score represents the harmonic mean of the precision and the recall. The higher the F1-score, the better the correspondence between predictions and ground truths is. The mean Average Precision depends on the individual Average Precisions (AP) obtained in each class (traffic light state). Basically, the AP of a class represents the area under the cumulative precision-recall curve~\cite{everingham2010pascal}. Then, mAP is calculated as the arithmetic mean of all APs. 

\added{The calculation of the F1-score is based on an optimal confidence threshold calculated over a validation set as described next. On the other hand, the mAP is calculated over all predictions, i.e., the confidence threshold is set to zero and no validation data is required. Finally, the mAP is also used as the selecting metric in the box-validation, as described last.}

\paragraph{\added{Validation}}
The validation sets are used to investigate the confidence threshold that yields the best F1-score result.
Thresholds from 0.01 to 1.0 in steps of 0.01 were considered, i.e., for each step, only predictions with confidence score equal or higher to the threshold were considered. Once the best confidence threshold was found, it was adopted as an optimal parameter of the final application, being used for calculating the corresponding F1-score for each test dataset.

\paragraph{\added{Box validation}} For the box-validation, the evaluation metrics were computed for different proportions of the prediction bounding boxes to confirm the effectiveness of the models. For each dataset's box-validation set, the metrics were calculated considering the areas of the prediction boxes multiplied by a factor $f = 0.4, 0.5, 0.6, \ldots, 1.9$, with the boxes' centers and aspect ratios preserved, so that it would be possible to find a factor which would make the prediction boxes fit best with the ground truths and, therefore, yield best evaluation results. Then, the metrics were calculated for the test sets considering the boxes proportion which yielded the highest mAP for the respective box-validation sets (adopting the value of $f$ which is closer to 1.0 as tiebreaker). It is worthy to emphasize that this procedure has a different semantic of a conventional validation step: the latter intends to find the optimal model for the task, whereas the box-validation focuses on compensating the inaccurate annotation which can mislead the performance assessment. To enable a fair evaluation, this procedure was repeated for each of the methods being compared, including the model trained with real data.

\subsection{Experiments}

The experiments aim at evaluating whether a traffic light detector can be trained without real data from the target domain, as well as investigating the influence of context on the learning process and the effectiveness of using the proposed method as data augmentation.

\subsubsection{Context impact analysis}
\label{sec:context_analysis}

These experiments aim to evaluate the deep detector's capability of learning from totally uncontextualized data, as well as to investigate the impact of the context in the detector performance. For this, the Fully Contextualized and Uncontextualized datasets were properly trained, validated, box-validated and tested to have their performance compared against each other.

\subsubsection{Templates \textnormal{versus} 3D models}

The impact of using traffic lights' 3D models was measured by comparing it with using simple 2D templates, as in~\cite{tabelini2019ijcnn}. For this, the test results yielded by a model trained, validated and box-validated with the Templates Only dataset are compared to the obtained with the Uncontextualized-based model.

\subsubsection{\added{Background domain analysis}}

\added{Torres \textit{et al.}~\cite{tabelini2019ijcnn} argue that using images from the problem domain as backgrounds may impair the detector performance, as the target-object may occur in the image and be also treated as background. To also confirm this issue in the traffic light application, experiments were conducted in order to compare the performance of models trained with backgrounds from the problem domain containing or not containing traffic lights (i.e., Positive Backgrounds and Negative Backgrounds datasets respectively) against their equivalent version trained with non-traffic-related backgrounds (i.e., the Uncontextualized dataset).}

\subsubsection{Use of the proposed method as data augmentation}

Experiments were also performed to evaluate the use of the proposed method as data augmentation, i.e., as an improvement for models based on real traffic scenes. The following experiments were conducted:

\begin{itemize}
    \item Context + Real-world: the Fully Contextualized training set was mixed with the Real-world Reference training set, creating a new set with 140k images. Then, a model was trained using this set as input during 140k iterations, so that each image was iterated once. Evaluation was performed not only on the final trained model but also on the checkpoint model corresponding to the first 70k iterations, since this was the number of iterations used to train the proposed method and the Real-world Reference;
    \item Fine tuning on real data: a 70k-iterations training on the Real-world Reference training set was performed to fine tune the model previously trained with the Fully Contextualized set.
\end{itemize}

For determination of the best confidence threshold to be used for testing these experiments, the Fully Contextualized validation set was used.

\subsubsection{Synthetic \textnormal{versus} real-world data}

To compare the results achieved with training using synthetic data against results for real-world data, the Real-world Reference was trained, validated, box-validated and tested so that the test results could be compared with the obtained with the model trained with the Fully Contextualized dataset. Exceptionally for the Real-world Reference, the best confidence threshold was not obtained through evaluation on a specific validation set, but obtained individually for each box-validation set. It is expected that this bias the results in favor of the Real-world reference model so that the performance of the Fully Contextualized-based model could be compared to the best possible case obtained with the reference for each dataset.

\subsection{Computational resources}

\removed{The preparation of the training datasets was processed on a Intel\textsuperscript{\textregistered} Core\textsuperscript{\texttrademark} i5-7200U CPU (2.50GHz, 8GB RAM), a Intel\textsuperscript{\textregistered} Xeon\textsuperscript{\textregistered} CPU X5690 (3.47GHz, 50GB RAM) and a Intel\textsuperscript{\textregistered} Xeon\textsuperscript{\textregistered} CPU E5606 (2.13GHz, 22GB RAM).}
The training and inference processes were performed on a Intel\textsuperscript{\textregistered} Core\textsuperscript{\texttrademark} i7-4770 CPU (3.40GHz, 16GB RAM) and an NVIDIA TITAN Xp GPU with 12GB memory, which performs one training iteration in approximately 0.35 seconds \added{and inference on a image in about 0.13 seconds}.
\section{Results and discussion}

\begin{figure*}[t]
    \begin{center}
        \includegraphics[width=\linewidth]{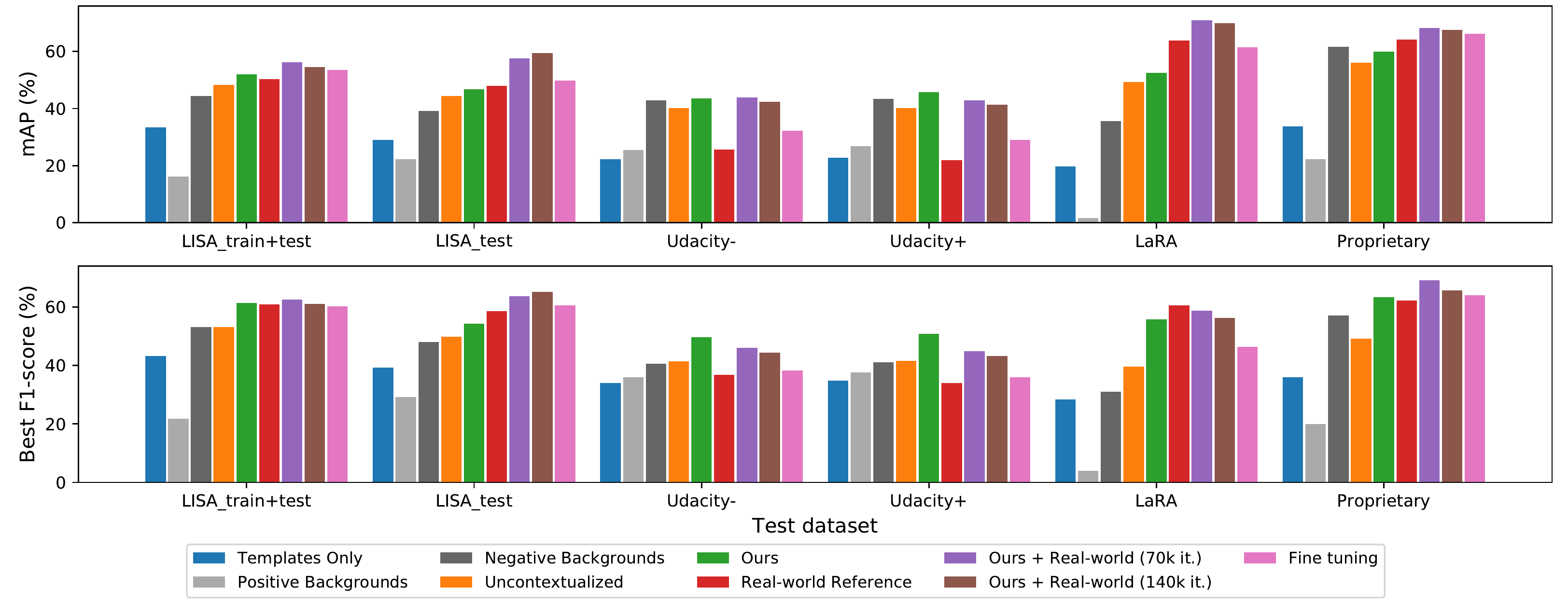}
    \end{center}
    \caption{mAP and best F1-score across the test datasets for all trained models.}
    \label{fig:test_results}
\end{figure*}

Figure~\ref{fig:test_results} shows the results of the mAP and F1-score obtained on the test datasets for all trained models, considering the value $f$ of the bounding boxes' area multiplier (one $f$ for each of the evaluated methods) which yielded the highest results on their respective box-validation sets.

\subsection{Context impact analysis}

The results in Figure~\ref{fig:test_results} reveal that the model trained with the Fully Contextualized dataset (Ours\added{; green bar}) performs better than the one trained with its counterpart (Uncontextualized\added{; orange bar}) for all test datasets and both evaluation metrics. In all test cases, the presence of context yields a mAP increasing within a range of approximately 2 to 6 p.p. The smallest gain occurs for LISA\_test (from 44.46 to 46.68\%) and the highest for Udacity+ (from 40.16 to 45.74\%). Meanwhile, the F1-score presents a significant increase for some test datasets. For the LaRA dataset, for example, the F1-score increases by more than 16 p.p. (from 39.60 to 55.78\%) with the insertion of context to the scene, while the Proprietary dataset presents an increasing of more than 14 p.p. (from 49.24 to 63.36\%). For the remaining datasets, the F1-score presents less pronounced increasing (minimum of 4.36 p.p, for LISA\_test, and maximum of 9.33 p.p., for Udacity+), but no decreasing to any case.

A direct analysis of the results of precision and recall, considering the confidence thresholds which yield the best F1-scores on the validation sets, reveals that the superior performance of the model trained with Fully Contextualized data is in almost all cases associated with significant increases in precision, while recall variation is less pronounced. For the Proprietary dataset, for example, the precision increases from 46.58 to 71.56\%, while the recall varies from 58.19 to 56.88\%. Overall, increasing in precision is around 10-25 p.p. depending on the dataset, while recall does not vary more than 2.5 p.p. The only exception stands for LISA\_test, which suffers a decrease in recall of approximately 9 p.p. The precision increasing is, however, higher than 15 p.p.

\subsection{Templates \textnormal{versus} 3D models}

Figure~\ref{fig:test_results} shows that the model trained with Uncontextualized data \added{(orange bars)} outperforms significantly the training with Templates Only \added{(blue bars)} in all scenarios. The differences in mAP range from 15.01, for LISA\_train+test, to 29.61 p.p., for LaRA, while in F1-score they range from 6.74, for Udacity+, to 13.26 p.p., for the Proprietary dataset.

This experiment shows that training using 3D models yields better results than training with 2D templates despite the similar appearance. In fact, the tridimensional aspect makes the traffic light models way more realistic and, therefore, provide better matching with real-world traffic lights.

\subsection{\added{Background domain analysis}}

\added{Overall, the results shown in Figure~\ref{fig:test_results} evidence that the performance of the model based on the Positive Backgrounds dataset (light-gray bars) are considerably inferior when compared to the Uncontextualized-based model (orange bars). The differences in mAP range from 13.42 (for Udacity+) to 47.73 p.p. (for LaRA) and the F1-score presents little difference for Udacity-based models only, while for the remaining datasets it ranges from 20.78 (for LISA\_test) to 35.63 p.p. (for LaRA). When analysed individually, the model reveals not to perform well. The mAP is limited to 26.74\% and the F1-scores are not higher than 37.62\% (both for Udacity+). On the other hand, the Negative Backgrounds model (dark-gray bars) seems to perform comparably with the Uncontextualized model. The most evident differences occur for LaRA and the Proprietary dataset. While Uncontextualized performance is superior by 13.69 p.p. in mAP and 8.56 p.p. in F1-score for the former, it is inferior by 5.46 p.p. in mAP and 7.90 p.p. in F1-score for the latter.}

\added{The results confirm that using domain-related backgrounds containing the target-object impairs performance, as stated in~\cite{tabelini2019ijcnn}. However, domain-related backgrounds without the target objects yields performance comparable to backgrounds from non-related domains. This motivates even further the use of the proposed method since it requires no real-world data from the problem domain.} 

\subsection{Use of the proposed method as data augmentation}

According to Figure~\ref{fig:test_results}, the comparison between the training with the Real-world Reference \added{(red bars)} and with mixed data (context + Real-world Reference; \added{purple and brown bars for 70k and 140k iterations, respectively}) reveals in general that using synthetic data as complement for real-world data improves or, at least, preserves performance, showing little decrease for LaRA's F1-score only.
Doubling the number of iterations (from 70 to 140k) kept the results of the augmentation nearly unaltered. The highest gain occurs for Udacity+, with more than +21 p.p. for mAP and +10 p.p. for F1-score. The worse performance occurs for the F1-scores of LISA\_test and LaRA. In its turn, fine tuning the context-based model with real-world data \added{(pink bars)} is less promising.
Its performance is considerably worse than mixing data for LISA\_test, Udacity-, Udacity+ and LaRA and comparable for the remaining datasets. Also, the performance comparison between the fine tuning and the Real-world Reference reveals that, for almost all cases, they perform comparably, indicating that the knowledge on real-world data may be strongly predominating over the synthetic data. 

When comparing the performance of the proposed method (Fully Contextualized; \added{green bars}) with the augmentation models, it is noticeable that in general the performance is also improved for all non-Udacity-based datasets when augmentation is applied, with a gain of up to 18.38 p.p. for mAP for the 70k-iterations data mixing (tested on LaRA) and 10.86 p.p. for F1-score for the 140k-iterations data mixing (tested on LISA\_test). Fine tuning tested in LaRA opposes to this in F1-score (46.37\% from fine tuning against 55.78\% from the Fully Contextualized model), but compensates in mAP (61.44 against 52.55\%). In addition, although augmentation does not provide improvements for Udacity compared to the Fully Contextualized model, the mixed data model with 70k iterations yield nearly the same mAP and little degradation for the F1-score, limited to 6.06 p.p. (Udacity+; Fully Contextualized against mixed data and 70k iterations).

Overall, the results show that the proposed method tends to be effective as data augmentation. Additionally, it is preferable to train models with both real-world and synthetic data, in which the deep detector learns both patterns simultaneously, than to train with synthetic data and then refine the learned pattern by fine tuning with real-world data.

\subsection{Synthetic \textnormal{versus} real-world data and discussion on the proposed method performance}

The model trained with the proposed method \added{(green bars)} outperforms the Real-world Reference \added{(red bars)} for Udacity-, Udacity+ and LISA\_train+test, although the negligible difference for the latter (1.7 p.p. of mAP and 0.56 p.p. of F1-score). For LISA\_test and LaRA, the real-world model outperforms the proposed method, but the highest difference is limited to 11.29 p.p. for mAP and 4.87 p.p. for F1-score, obtained for LaRA.

Considering the results obtained for all test datasets, the proposed method achieves an average mAP of 50.08\% $\pm$ 5.99\% and an average F1-score of 55.93\% $\pm$ 5.50\%, while the Real-world Reference yields average mAP and F1-score of, respectively, 45.61\% $\pm$ 18.26\% and 52.21\% $\pm$ 13.1\%. \camreadyadd{
Overall, these results are comparable to recent deep learning approaches for traffic light detection ~\cite{possatti2019traffic,pon2018hierarchical,kim2018efficient}. They report mAP ranging from 38 to 55\% in a intra-database scenario, which tends to be less challenging than the investigated cross-database scenario. Despite its relevance, the latter scenario is usually overlooked in the literature and, in this work, it is covered by the Real-world Reference. Given the results obtained with both Ours and the Real-world Reference models, it is evidenced that the proposed method achieves performance competitive to the state-of-the-art methods and the choice of the real-world baseline was satisfactory even with the intrinsic challenges of the cross-database approach.} The results suggest that the proposed method has the potential to match with a model trained with real-world data without manual effort for acquisition and annotation of data. Moreover, the results were further enhanced by mixing synthetic and real-world data, yielding an average mAP of 56.63\% $\pm$ 11.77\% and an average F1-score of 57.55\% $\pm$ 9.93\% for the 70k-iterations model.

\begin{figure}
\begin{center}
   \includegraphics[width=\columnwidth]{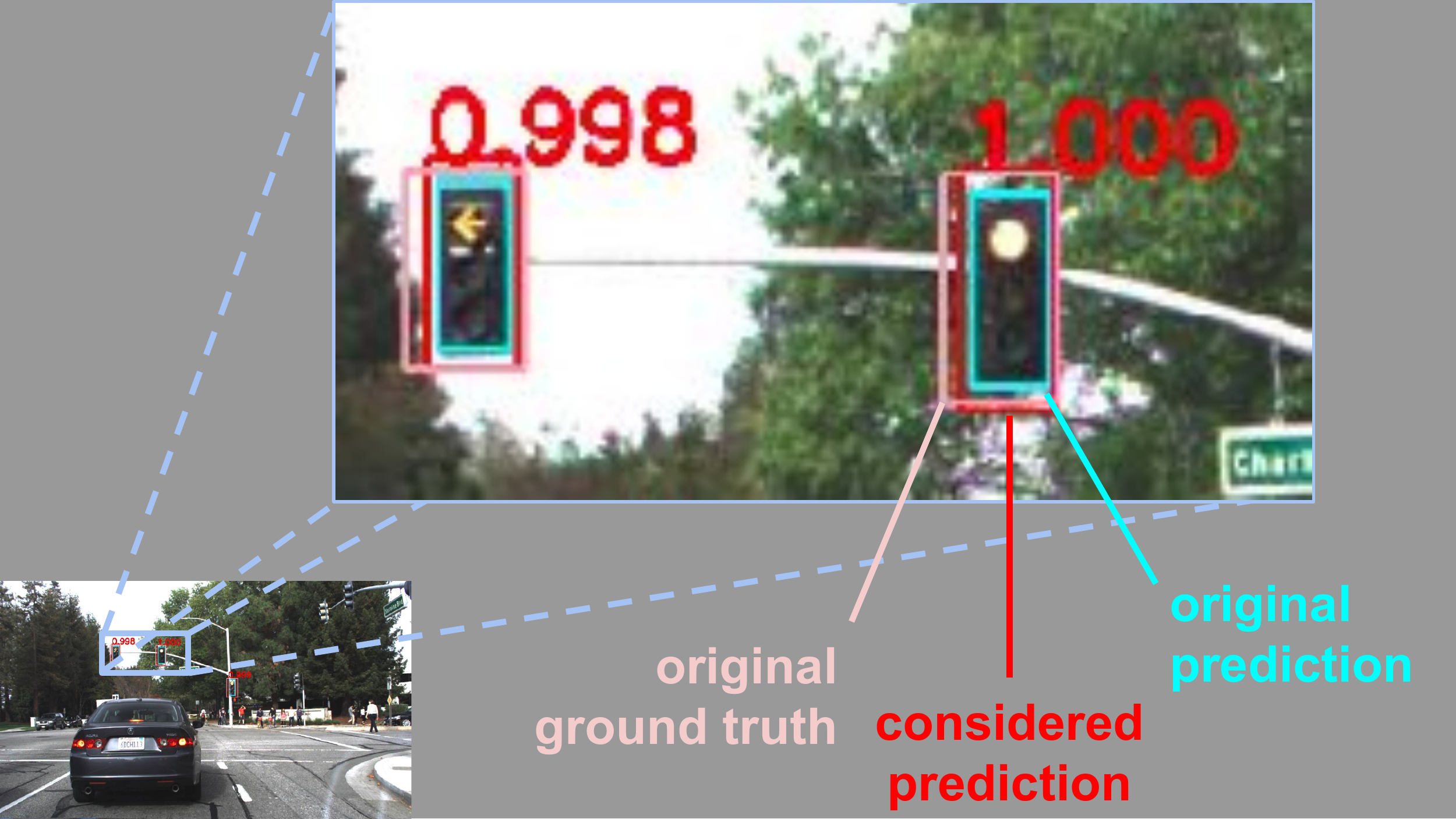}
\end{center}
   \caption{Example of inference result for an image from Udacity dataset. The light red boxes represent the original ground truth boxes. The cyan boxes correspond to the original prediction. The red boxes followed by the confidence score represents the adjusted predictions after box-validation.}
\label{fig:predictions}
\end{figure}

Figure~\ref{fig:predictions} shows a visual example of the proposed method results. The light red boxes represent the original ground truth boxes. The cyan boxes correspond to the original prediction, which fits best with the real traffic light face area. The red boxes followed by the confidence score represents the prediction bounding boxes with area multiplied by the optimal factor, making the prediction boxes' areas closer to the ground truth.

\section{Conclusion}
Detecting and recognizing traffic elements, particularly traffic lights, is essential for autonomous vehicles to drive safely and according to the traffic legislation. Although deep detectors are effective solutions for traffic light detection, acquisition and annotation of training data demand significant time and effort. Also, real data is subject to imbalance since yellow-stated traffic lights are less likely to be recorded than red and green. The proposed method tackles these issues by generating artificial data that simulate simple traffic context.

The results show that it is possible to obtain reasonable performance by just inserting artificial traffic lights in natural non-traffic-related contexts. Results become even better when a simple traffic context is modeled and added to the scene. The experiments showed that this proposal yields average mAP and average F1-score of approximately 50\% and 56\%, respectively, each nearly 4 p.p. higher than the respective results obtained by training with real-world traffic data. 

It is also clear that training with datasets built with little generation effort (related to the construction of the traffic scene) and no annotation efforts provides results comparable to results obtained with real training data exhaustively annotated. Although there are some traffic light datasets available, the application of this principle may enhance the performance for cases in which it is necessary to detect traffic light models distinct from the ones comprised by the available datasets. For example, models with more than three bulbs and/or in horizontal orientation may be added. Such flexibility enables the application of the detector in cities with different traffic light pattern without having to acquire and annotate a new dataset.

\section*{Acknowledgments}
\camreadyadd{This study was financed in part by the Coordenação de Aperfeiçoamento de Pessoal de Nível Superior - Brasil (CAPES) - Finance Code 001, Conselho Nacional de Desenvolvimento Científico e Tecnológico (CNPq, Brazil), PIIC UFES and Fundação de Amparo à Pesquisa do Espírito Santo - Brasil (FAPES) – grant 84412844. The authors thank NVIDIA Corporation for the donation of the GPUs used in this research.}

\bibliographystyle{cag-num-names}
\bibliography{references}



\end{document}